\documentclass[10pt,twocolumn,letterpaper]{article}

\usepackage{cvpr}
\usepackage{times}
\usepackage{epsfig}
\usepackage{graphicx}
\usepackage{amsmath}
\usepackage{amssymb}
\usepackage{comment}
\usepackage{bm}
\usepackage{here}

\usepackage[pagebackref=true,breaklinks=true,letterpaper=true,colorlinks,bookmarks=false]{hyperref}

\cvprfinalcopy 

\def\cvprPaperID{****} 

\ifcvprfinal\fi
\begin{document}

\setlength{\baselineskip}{11pt} 

\title{Interpretation of Feature Space using Multi-Channel Attentional Sub-Networks}

\author{Masanari Kimura $\dagger$, Masayuki Tanaka $\dagger, \ddagger$ \\
$\dagger$ National Institute of Advanced Industrial Science and Technology\\
$\ddagger$Tokyo Institute of Technology \\
}

\maketitle

\begin{abstract}
Convolutional Neural Networks have achieved impressive results in various tasks, but interpreting the internal mechanism is a challenging problem. To tackle this problem, we exploit a multi-channel attention mechanism in feature space. Our network architecture allows us to obtain an attention mask for each feature while existing CNN visualization methods provide only a common attention mask for all features.
We apply the proposed multi-channel attention mechanism to multi-attribute recognition task. We can obtain different attention mask for each feature and for each attribute. Those analyses give us deeper insight into the feature space of CNNs.
The experimental results for the benchmark dataset show that the proposed method gives high interpretability to humans while accurately grasping the attributes of the data.
\end{abstract}

\section{Introduction}

In recent years, Convolutional Neural Networks (CNNs) have made great achievements in various tasks \cite{krizhevsky2012imagenet, karpathy2014large}.
Despite such success, it is known that an interpretation of the CNNs is difficult for humans.
There are various kinds of research to understand the inference mechanism of CNNs to tackle this problem \cite{selvaraju2017grad, zhang2018visual, kuwajima2019improving}. In particular, "Visual explanation", which visualizes the inference mechanism of the CNN, is an important task.

We aim to obtain highly interpretable neural networks using an attention mechanism to acquire features that are important for classifiers. 
Our main idea is to train sub-networks with multi-channel attention mask for each attribute. The attention mask applied to the sub-network is not common in the feature map but has the same number of channels as the feature map. This multi-channel attention mechanism can reveal which channel in the feature map focuses on which part of the image.

\begin{figure}[htb]
  \centering
  \includegraphics[scale=0.4]{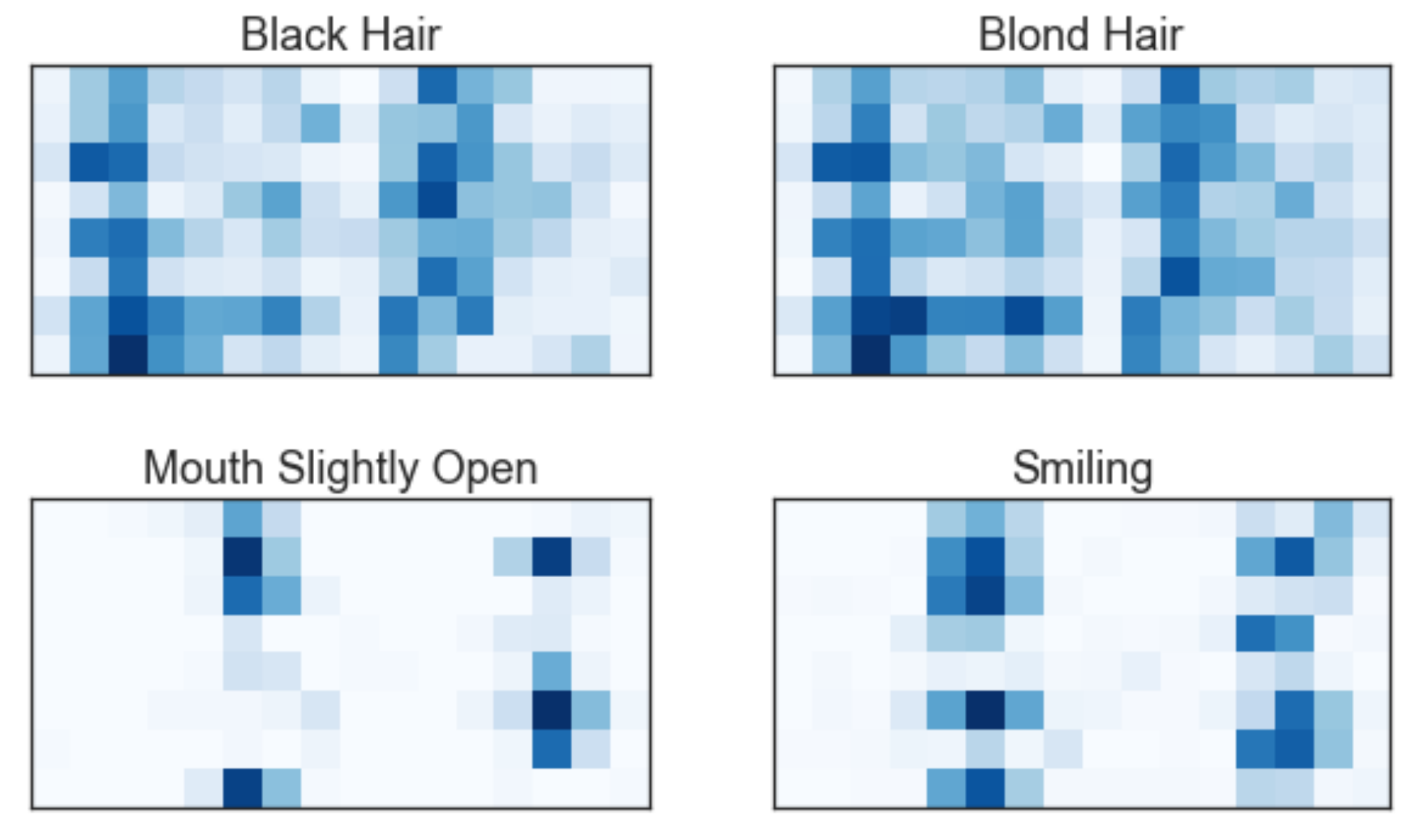}
  \caption{Average of attention mask of each feature. One element means one-dimensional mean value of attention mask, and there are 128 elements.}
  \label{fig:feature_vis}
\end{figure}

\begin{figure*}[htb]
  \centering
  \includegraphics[scale=0.47]{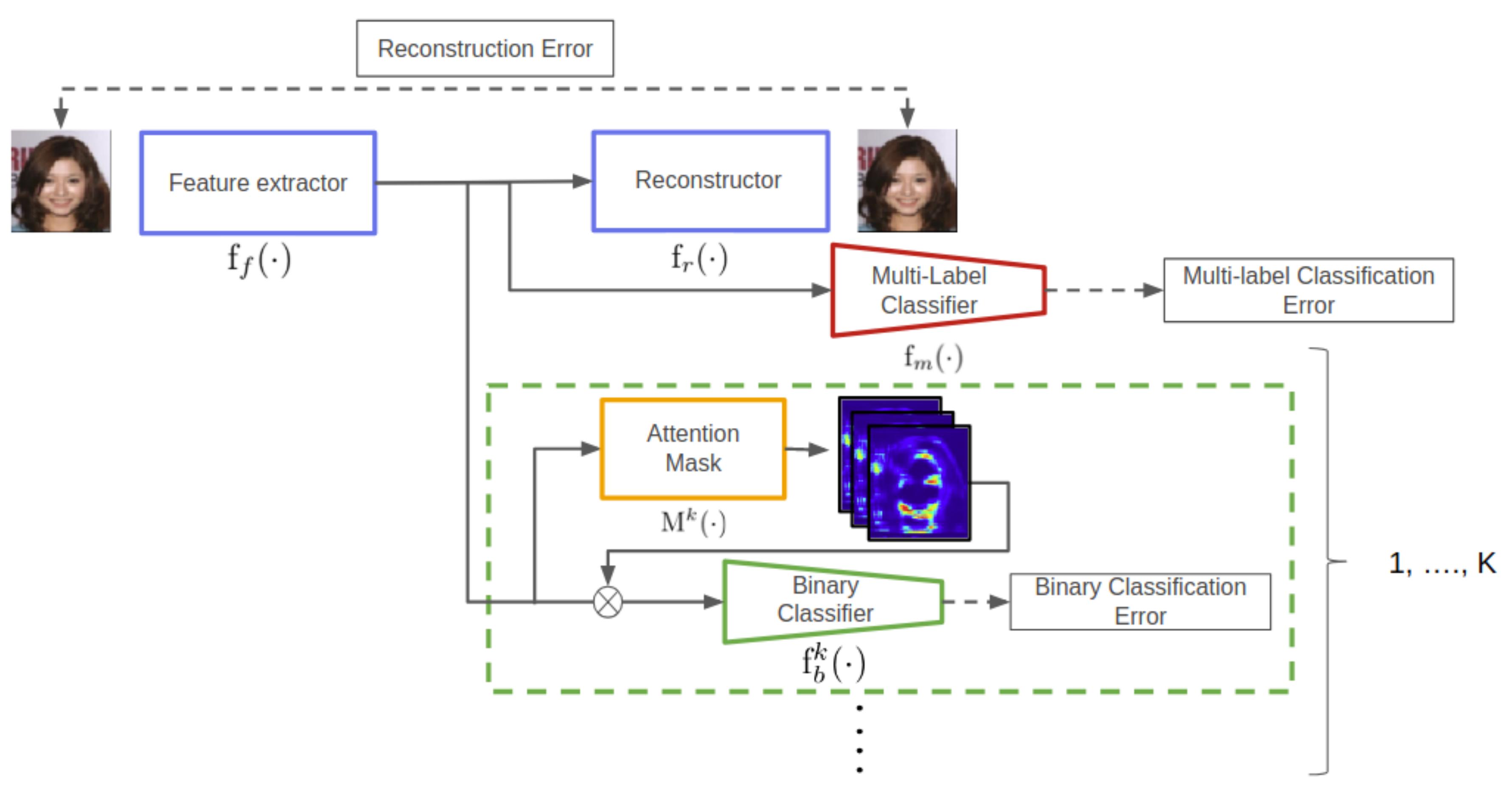}
  \caption{Overview of proposed network architecture. Here, there are $K$ binary classification components, where $K$ is the number of attributes.}
  \label{fig:network_structure1}
\end{figure*}

Figure \ref{fig:feature_vis} shows a grid visualization of the mean values of each channel of attention masks. Our analysis reveals that similar channels of features are commonly used for attributes of "Black Hair" and "Blond Hair". Important channels for attributes of "Mouth Slightly Open" and "Smiling" are also similar. This result suggests that the features used for each attribute are limited and can be applied to a very wide range of applications. 
Our contributions are as follows:
\begin{itemize}
    \item{We suggest that applying different attention masks to each channel of the feature map which gives us enables deeper insights into CNNs.}
    \item{We propose a novel framework to visualize important features for each attribute by using an attention mechanism.}
    \item{We performed a useful analysis on the features of CNNs using the proposed framework.}
\end{itemize}
Our analysis is highly versatile and leads to a broad range of applied research, such as improvement of classification accuracy, network pruning, image generation, and other applications.

\section{Related Works}
There are several methods for visualizing important features for CNNs in images \cite{selvaraju2017grad, zhang2018visual}.
Grad-CAM \cite{selvaraju2017grad} visualizes the area that brings a large gradient to the output neuron when the target class is specified. GAIN \cite{zhang2018visual} is the framework that provides direct guidance on the attention mask generated by weakly supervised learning. In addition, ABN \cite{fukui2018attention} made reasoning to CNN using the mechanism of Global Average Pooling (GAP).
WiG \cite{tanaka2018weighted} achieved the performance improvement of CNNs by applying the gating by attention mechanism as the activation function.

In our network architecture, we can obtain an attention mask for each feature while existing CNN visualization methods provide only a common attention mask for all features.

\begin{figure*}[htb]
  \centering
  \includegraphics[scale=0.39]{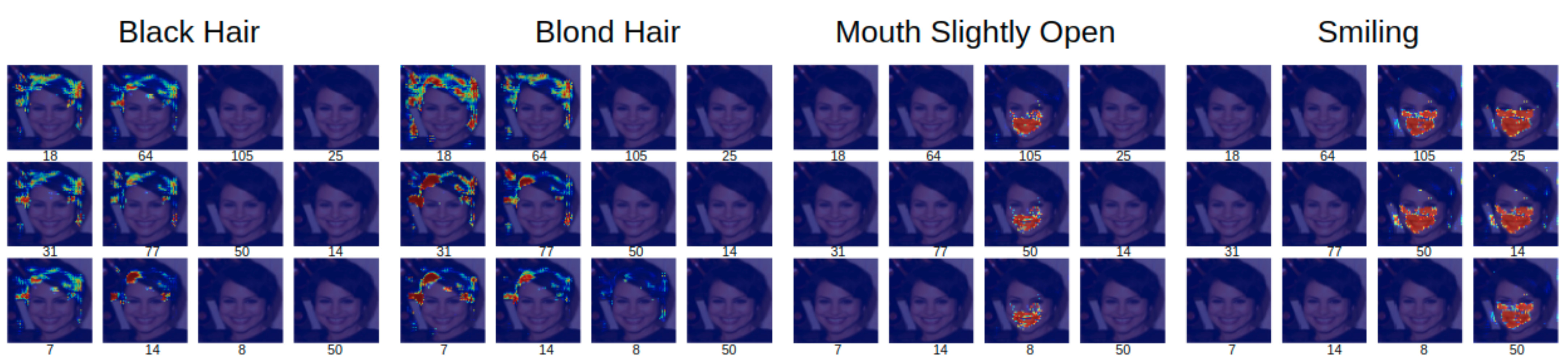}
  \caption{Visualizing attention masks on multiple facial attributes recognition. One element is one channel of the attention mask. The number under each mask means feature IDs.}
  \label{fig:attributes}
\end{figure*}

\section{Proposed Method}
We aim to interpret and theorize the internal mechanisms of CNNs using attention mechanism.
Figure \ref{fig:network_structure1} shows an overview of our proposed network architecture. In the proposed method, multiple outputs corresponding to an image with multiple attributes.

Let ${\bm X} = \{{\bm x}_1, {\bm x}_2,\dots,{\bm x}_n\}^N_{i=1}$ be a sample set, ${\bm Y} = \{{\bm y}_1, {\bm y}_2,\dots,{\bm y}_n\}^N_{i=1}$ be a label set, $N$ is the number of samples.
The details of each component are as follows.

\subsection{Feature Extractor}
Feature extractor $f_f$ extracts generic feature used by all the following networks. This component performs feature extraction. For this component, we use the Dilation Network \cite{fisher2016multi}.

\subsection{Binary Classifiers with Attention Mechanism}
Binary classifiers $F_b = \{f_b^1, f_b^2,\dots,f_b^k \}^K_{k=1}$, where $K$ is number of attributes, are components that perform binary classification corresponding to each attribute of the image.
This network is our main component.
The loss for the binary classifiers can be expressed by a sum of binary cross entropy of each attribute as
\begin{eqnarray}
    L_{b} &=& - \frac{1}{N} \sum^N_{i=1} \sum^K_{k=1} 
    y_{i}^{k} \log{\hat{y}_{b,i}^{k}}
    + (1 - y_{i}^k)\log{(1 - \hat{y}_{b,i}^{k})} \,, 
    \nonumber \\ \\
    \hat{y}_{b,i}^{k} &=& f_b^k((1+M^k(f_f({\bm x}_i))) \otimes f_f({\bm x}_i))\,,
\end{eqnarray}
where $M^k(\cdot)$ is the attention mask for $k$-th attribute, and $\otimes$ represents element-wise product operation.
Note that the attention mask $M^k$ has the same number of channels as that of feature $f_f$.
It differentiates from existing importance visualization algorithms.
We apply attention mask as $1 + M^k(\cdot)$ to emphasize the area of interest while keeping the low value of the attention mask following in \cite{fukui2018attention}.

\subsection{Multi-Label Classifier}
Multi-label classifier is a component that classifies multiple labels.
The loss for the multi-label classifier is
\begin{eqnarray}
    L_{m} &=& - \frac{1}{N} \sum^N_{i=1} \sum^{K}_{k=1}  y_{i}^{k} \log{\hat{y}_{m,i}^{k}}
    + (1 - y_{i}^k)\log{(1 - \hat{y}_{m,i}^{k})} \,, 
    \nonumber \\ \\
    \hat{y}_{m,i}^k &=& f_m^k(f_f({\bm x}_i)) \,.
\end{eqnarray}
We put this network component to obtain better feature representation.

\subsection{Reconstructor}
Reconstructor $f_r$ is a component that reconstructs a input image from extracted feature. The reconstruction loss $L_r$ is as follows.
\begin{equation}
    L_r = \frac{1}{N} \sum^N_{i=1} ({\bm x}_i - f_r(f_f({\bm x}_i)))^2 \,.
\end{equation}
This component aims to obtain better feature representation $f_f(\cdot)$.

\subsection{Overall Loss Function}
The overall loss function is:
\begin{equation}
    L = \alpha \cdot L_b + \beta \cdot L_m + \gamma \cdot L_r  + \lambda \cdot \|{\bm M}\|_1\,,
\end{equation}
where, $\alpha$, $\beta$, $\gamma$, and $\lambda$ are weight parameters of each component,
and $\|{\bm M}\|_1$ means L1 sparseness to the attention mask which is used to extract features that are really important for the data.

\section{Experimental Results}
We evaluate our method using the CelebA dataset \cite{liu2015deep}, which consists of 40 facial attribute labels and 202,599 images (182,637 training images and 19,962 testing images).
The parameters of the proposed method are $\alpha = 1$, $\beta = 1$, $\gamma = 4$ and $\lambda = 0.00001$. The dimension of the attention mask and feature map is 128.
The reproduction code is available online\footnote{http://www.ok.sc.e.titech.ac.jp/\%7Emtanaka/proj/mam/}.

Figure \ref{fig:attributes} shows the visualization of the attention mask by our proposed method. In this figure, common feature channels are visualized for each attribute, and each column means the top three features which have high importance for each attribute. Our attention masks focus on areas that may be important to attributes. In addition, this experimental result suggests that analysis of feature space reveals the relationship among attributes. For example, feature IDs 25, 14 and 50 are not used in Mouse Slightly Open, although they are used in Smiling. On the other hand, IDs 105 and 50 are used in the therapy of Smiling and Mouse Slightly Open, and ID 8 is not used in Smiling.
Only the mouth region is focused for the attribute of Mouse Slightly Open, while a wide region including mouse and eyes are focused for the attribute of Smiling.
Those results are consistent with the human instinct.

\begin{figure}[htb]
  \centering
  \includegraphics[scale=0.38]{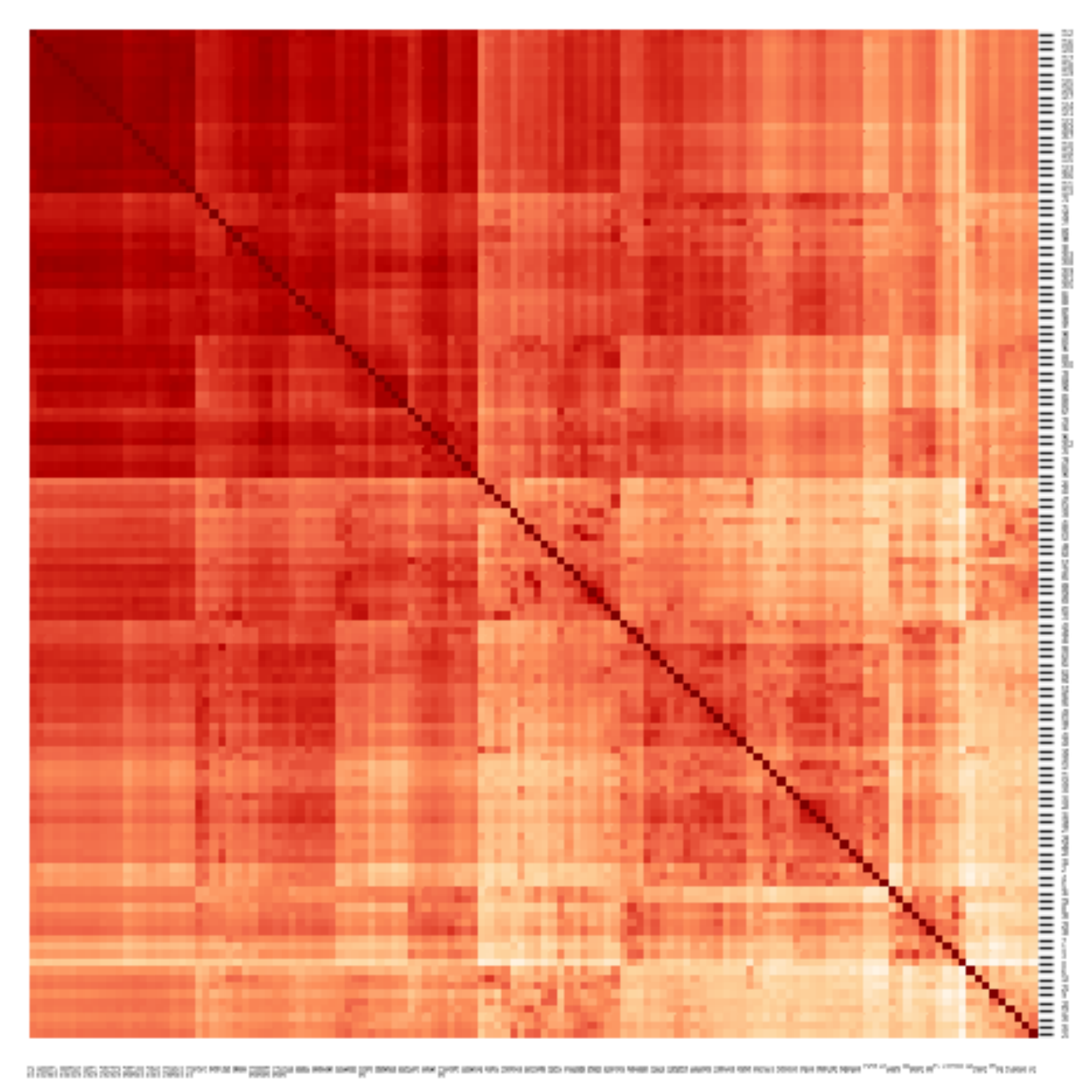}
  \caption{Visualization of correlation in feature space. In this figure, deeper color means higher correlation.}
  \label{fig:cor_matrix}
\end{figure}

Figure \ref{fig:cor_matrix} shows the visualization of the correlation of the feature space. 
Table \ref{table:feature_correlation_top5} lists some of the feature IDs and their highly correlated features.
Our multi-channel attention mechanism makes it possible to obtain correlations among each channel of the feature map.

Table \ref{table:correlation_top5} lists some of the attributes and their highly correlated attributes. Correlations of attributes are estimated based on correlations of features.
Attributes that are intuitively similar are highly correlated. This result makes it possible to group highly correlated attributes. In addition, experimental results may even reveal potential relationships among attributes.

Table \ref{table:accuracy_celeba} shows the experimental results of the classification task in the CelebA dataset.
In this experiment, MT-RBM PCA \cite{ehrlich2016facial}, LNets+ANet \cite{liu2015deep}, and FaceTracer \cite{kumar2008facetracer} are used as comparison methods.
The proposed method achieves good performance with many attributes and all average accuracy.

\begin{table*}[ht]
\centering
\small
\caption{Correlation among the features. It lists the target features, highly correlated features with the target, and correlation.}
\label{table:feature_correlation_top5}
\begin{tabular}{r|r|r|r|r|r}
\hline
Target Feature ID & \multicolumn{5}{c}{Top5 Highly Correlated Feature IDs}                      \\ \hline
1              & 72 (0.98)  & 87 (0.95)  & 44 (0.94)  & 92 (0.94)  & 87 (0.94)  \\
32             & 111 (0.96) & 114 (0.95) & 100 (0.95) & 15 (0.94)  & 119 (0.94) \\
64             & 127 (0.97) & 15 (0.97)  & 119 (0.97) & 114 (0.97) & 57 (0.97) \\ \hline
\end{tabular}
\end{table*}

\begin{table*}[ht]
\centering
\small
\caption{Correlation among the attributes. It lists the target attributes and the top five attributes that are highly correlated with the target. }
\label{table:correlation_top5}
\begin{tabular}{l|l}
\hline
Target  Attribute & Top5 Highly Correlated Attributes                                                        \\ \hline
Black Hair        & Blond Hair, Brown Hair, Bald, Wearing Hat, Gray Hair                   \\
Heavy Makeup      & Wearing Lipstick, Male, Rosy Cheeks, Attractive, Young                 \\
Bushy Eyebrows    & Bags Under Eyes, Eyeglasses, Arched Eyebrows, Heavy Makeup, Attractive \\
\hline
\end{tabular}
\end{table*}

\begin{table}[h]
\centering
\vspace{0mm}
\caption{Classification accuracy on the CelebA dataset. In this experiment, MT-RBM PCA \cite{ehrlich2016facial}, LNets+ANet \cite{liu2015deep}, and FaceTracer \cite{kumar2008facetracer} are used as comparison methods.}
\label{table:accuracy_celeba}
\footnotesize
\begin{tabular}{l|lllll}
\hline
Attribute              & Ours        & \cite{ehrlich2016facial} & \cite{liu2015deep} & \cite{kumar2008facetracer} \\ \hline
5 Shadow               & {\bf 92.85} & 90                       & 91                 & 85                 \\
Arched Eyebrows        & {\bf 81.37} & 77                       & 79                 & 76                 \\
Attractive             & 80.71       & 76                       & {\bf 81}           & 78                 \\
Bags Under Eyes        & {\bf 83.79} & 81                       & 79                 & 76                 \\
Bald                   & {\bf 98.30} & 98                       & 98                 & 89                 \\
Bangs                  & 94.10       & 88                       & {\bf 95}           & 88                 \\
Big Lips               & {\bf 70.14} & 69                       & 68                 & 64                 \\
Big Nose               & {\bf 83.67} & 81                       & 78                 & 74                 \\
Black Hair             & {\bf 88.39} & 76                       & 88                 & 70                 \\
Blond Hair             & {\bf 95.10} & 91                       & 95                 & 80                 \\
Blurry                 & {\bf 95.33} & 95                       & 84                 & 81                 \\
Brown Hair             & {\bf 86.55} & 83                       & 80                 & 60                 \\
Bushy Eyebrows         & {\bf 91.87} & 88                       & 90                 & 80                 \\
Chubby                 & {\bf 96.02} & 95                       & 91                 & 86                 \\
Double Chin            & {\bf 96.68} & 96                       & 92                 & 88                 \\
Eyeglasses             & 98.67       & 96                       & {\bf 99}           & 98                 \\
Goatee                 & {\bf 96.72} & 96                       & 95                 & 93                 \\
Gray Hair              & {\bf 97.89} & 97                       & 97                 & 90                 \\
Heavy Makeup           & 89.49       & 85                       & {\bf 90}           & 85                 \\
High Cheekbone         & 86.77       & 83                       & {\bf 87}           & 84                 \\
Male                   & 97.38       & 90                       & {\bf 98}           & 91                 \\
Mouth Open             & {\bf 93.67} & 82                       & 92                 & 87                 \\
Mustache               & 96.60       & {\bf 97}                 & 95                 & 91                 \\
Narrow Eyes            & {\bf 86.38} & 86                       & 81                 & 82                 \\
No Beard               & 94.87       & 90                       & {\bf 95}           & 90                 \\
Oval Face              & {\bf 73.33} & 73                       & 66                 & 64                 \\
Pale Skin              & {\bf 97.67} & 96                       & 91                 & 83                 \\
Pointy Nose            & {\bf 75.62} & 73                       & 72                 & 68                 \\
Recede Hair            & 93.44       & {\bf 96}                 & 89                 & 76                 \\
Rosy Cheeks            & {\bf 94.67} & 94                       & 90                 & 84                 \\
Sideburns              & {\bf 97.65} & 96                       & 96                 & 94                 \\
Smiling                & {\bf 92.28} & 88                       & 92                 & 89                 \\
Straight Hair          & {\bf 81.60} & 80                       & 73                 & 63                 \\
Wavy Hair              & {\bf 81.64} & 72                       & 80                 & 73                 \\
Earring                & {\bf 84.61} & 81                       & 82                 & 73                 \\
Hat                    & 98.92       & 97                       & {\bf 99}           & 89                 \\
Lipstick               & 92.52       & 89                       & {\bf 93}           & 89                 \\
Necklace               & 86.37       & {\bf 87}                 & 71                 & 68                 \\
Necktie                & {\bf 96.30} & 94                       & 93                 & 86                 \\
Young                  & {\bf 87.00} & 81                       & {\bf 87}           & 80                 \\ \hline
Average                & {\bf 92.05} & 87                       & 87                  & 81　\\
\hline
\end{tabular}
\end{table}

\section{Conclusion and Discussion}
We proposed a novel network architecture and attention mechanism that can give a visual explanation of CNNs. 
Our multi-channel attention mechanism makes it possible to obtain correlations among each channel of the feature map. We suggest that analysis of feature maps obtained by the proposed method is highly versatile and lead to a broad range of applied research, such as improvement of classification accuracy, network pruning, image generation, and other applications.

\section{Acknowledgements}
{\small
A part of this paper is based on results obtained from a project commissioned by the New Energy and Industrial Technology Development Organization (NEDO).
}

{\footnotesize
\bibliographystyle{ieee_fullname}

}

\end{document}